\documentclass[acmsmall,screen]{acmart}
\setcopyright{none}
\renewcommand\footnotetextcopyrightpermission[1]{} 
\usepackage{enumitem}
\usepackage{amsmath,amsfonts}
\usepackage{graphicx}
\usepackage{textcomp}
\usepackage{xcolor}
\usepackage{hyperref}
\usepackage{tikz}
\usepackage{subfiles}
\usepackage{subcaption}
\usepackage{algpseudocode} 
\usepackage{siunitx,booktabs} 

\usepackage{color,soul} 
\usepackage{flushend}
\usepackage{tikz,pgfplots}
\newcommand{\hypvec}[1]{\vec{\mathcal{#1}}}
\usepackage{pifont}
\newcommand{\cmark}{\ding{51}}%
\newcommand{\xmark}{\ding{55}}%
\usetikzlibrary{patterns}
\usepackage{booktabs}
\usepackage{multirow}
\usepackage{amsmath,graphicx,color}
\usepackage{rotating}
\usepackage{array}
\usepackage{graphicx}
\usepackage{tabu}
\usepackage{booktabs}
\newcommand\invcircledast[1]{%
  \tikz[baseline=(X.base)] 
    \node (X) [draw, shape=circle, inner sep=0, fill=black, text=white] {\strut \small \textbf{#1}};%
}

\usepackage{array}
\usepackage{caption}
\usepackage{graphicx}
\usepackage{tabu}
\usepackage{multirow}
\usepackage{ragged2e}
\usepackage{booktabs}
\pgfplotsset{compat=1.11}

\newcommand{\Design}[0]{the framework }

\begin{document}
\title{Efficient Personalized Learning for Wearable Health Applications using HyperDimensional Computing}
\pagestyle{plain}

 \author{Sina Shahhosseini, Yang Ni, Hamidreza Alikhani, Emad Kasaeyan Naeini, Mohsen Imani, Nikil Dutt, Amir M. Rahmani}
 \affiliation{
 \institution{University of California Irvine}
 \country{USA}
 }
 \email{{sshahhos, yni, hamidra, ekasaeya, m.imani, dutt, a.rahmani}@uci.edu}
\begin{abstract}
Health monitoring applications increasingly rely on machine learning techniques to learn end-user physiological and behavioral patterns in everyday settings. Considering the significant role of wearable devices in monitoring human body parameters, on-device learning can be utilized to build personalized models for behavioral and physiological patterns, and provide data privacy for users at the same time. However, resource constraints on most of these wearable devices prevents the ability to perform online learning on them. To address this issue, it is required to rethink the machine learning models from the algorithmic perspective to be suitable to run on wearable devices. Hyperdimensional computing (HDC) offers a well-suited on-device learning solution for resource-constrained devices and provides support for privacy-preserving personalization. Our HDC-based method offers flexibility, high efficiency, resilience, and performance while enabling on-device personalization and privacy protection. We evaluate the efficacy of our approach using three case studies and show that our system improves energy efficiency of training by up to $45.8\times$ compared with the state-of-the-art Deep Neural Network (DNN) algorithms while offering a comparable accuracy.
\end{abstract}
\maketitle
\section{Introduction}
Wearable devices play a significant role in health monitoring systems by continuously monitoring  human physiological and physical data~\cite{sazonov2020wearable}. Such signals are acquired to gain insight into health trends and provide actionable information to patients. 
Health and wellness applications increasingly rely on 
machine learning (ML) algorithms to capture the user's behavioral and physiological patterns~\cite{panch2018artificial},
but pose two challenges: 
1) inability to perform on-device learning for resource-constrained wearables, as ML workloads require significant computational power and storage~\cite{brandalero2020multi, sapra2020constrained}, and
2) the need to develop on-device and online learning algorithms that support privacy-preserving personalization. Nowadays, the majority of wearable devices (e.g., smart watches) are multi-application and capable of interacting with users to collect feedback and personalize their models over time to meet the unique characteristics of each person, however, online training of ML models on wearable devices currently is not feasible using state-of-the-art machine learning algorithms ~\cite{dhar2019device}. 
For these reasons, alternative learning algorithms are required to deliver real-time, low-power, and personalize services on wearable devices, and therefore, it is desired to re-design the learning process with respect to both algorithm and hardware. 


Hyperdimensional computing (HDC) offers an alternative computational model mimicking \textit{"the human brain"} in its functionality~\cite{kanerva2009hyperdimensional}. 
HDC is based on the understanding that the human brain operates on \textit{high-dimensional} representations. 
It maps data points into a high-dimensional space to learn a model with near-linear training time. 
HDC offers a well-suited solution for online learning and personalization on resource-constrained devices since: 
(i) HDC models are computationally efficient, highly parallel to train, and amenable to hardware-level optimization~\cite{khaleghi2021tiny,imani2017exploring}. 
(ii) HDC can naturally enable on-device online learning for wearable devices~\cite{hernandez2021onlinehd}, thereby facilitating privacy-preservation and personalization,  and (iii) HDC offers a robust solution to noise~\cite{salamat2020accelerating}. 
There have been recent efforts~\cite{rahimi2017hyperdimensional,rahimi2020hyperdimensional,benatti2019online,burrello2020ensemble,moin2021wearable} 
to deploy HDC algorithms to offer efficient on-device learning using single-pass training. However, single-pass training provides very weak classification accuracy compared to online learning in HDC~\cite{hernandez2021onlinehd}.

Furthermore, recent wearable solutions mainly focus on monitoring vital signs, which has limitations for wellness prediction. 
The data obtained from various signal sources might span multiple dimensions across multiple scales and exhibit varying precision. 
Some features may emerge due to users' behavioral patterns and context, which will impose a much higher degree of variation from one user to another user. 
These variations result in degrading the performance of general wearable health applications. Thus, it is imperative to integrate personalized health-related data collected from various sources. 
Existing on-device learning solutions~\cite{rahimi2020hyperdimensional,moin2021wearable,benatti2019online,burrello2020ensemble,bhat2019ultra} fail to offer personalized learning for wearable devices. 
In addition, these solutions are targeted for specific applications and platforms limiting their utility for multi-purpose wearable devices. For example, Moin et al.~\cite{moin2021wearable} proposed a custom hand gesture recognition system, while Bhat et al.~\cite{bhat2019ultra} proposed a DNN-based activity recognition system based on Application Specific Integrated Circuit (ASIC) platform, making the solutions application-specific.

Our approach enables accurate online on-device training and avoids model saturation by adopting and customizing the HDC training strategy presented in \cite{hernandez2021onlinehd} for resource-constrained wearable devices. 
A preliminary version of the approach has been proposed in ~\cite{shahhosseini2022flexible} with the following contributions:
\begin{itemize}[leftmargin=*]
    \item We implemented an online training framework for fast and accurate learning using HDC algorithms on both CPU and FPGA platforms. We presented the HDC framework as a practical and flexible solution for efficient on-device and online learning for wearable applications. We also demonstrate that on-device learning enables \textit{personalization} and \textit{user privacy protection} for wearable devices. The CPU implementation of our approach can be readily implemented on existing off-the-shelf multi-purpose wearable devices such as smartwatches.
    \item We demonstrated the effectiveness of our solution using three case studies compared with SOTA learning algorithms. Our evaluation showed that our HDC-based system improves training performance of the wearables by up to $35.8\times$ compared with DNN while providing comparable accuracy.    
\end{itemize}

In addition to promising results obtained in terms of accuracy and performance, another important superiority of HDC compared to other methods like DNN is its immense energy efficiency gain. This gain emanates from utilizing simple and low energy bit-wise operations and avoiding compute-intensive gradient operations in HDC. In this paper, we propose an efficient, flexible, and personalized HDC-based learning approach for wearable devices running health applications. Besides, HDC relies on high-dimensional space for representation of data; Instead of floating-point operations in conventional learning algorithms, bit-wise operation is the basic of HD computing. Consequently, there is an inherent accuracy toleration against hardware errors. In this paper, we present robustness of the proposed framework against hardware errors. We summarize the contributions as follows:
\begin{itemize}[leftmargin=*]
    \item We demonstrate the efficiency of the HDC-based framework. Our evaluation shows that the proposed framework improves training and inference energy efficiency of the wearable devices by up to $45.8\times$ and $5.1\times$ compared with DNN, respectively while providing comparable accuracy to DNNs. 
    \item We present the robustness of the HDC-based framework against hardware errors. Our evaluation shows the HDC-based framework is up to $60.1\times$ more robust than the DNN algorithm.
\end{itemize}
\section{Related Work and Motivation}
\begin{table*}
\centering
\caption{Summary of on-device solutions for wearable health applications.}
\resizebox{1\linewidth}{!}{%
\begin{tabu}{>{\centering\hspace{0pt}}m{0.2\linewidth}>{\centering\hspace{0pt}}m{0.15\linewidth}>{\centering\hspace{0pt}}m{0.185\linewidth}>{\centering\hspace{0pt}}m{0.2\linewidth}>{\centering\hspace{0pt}}m{0.120\linewidth}>{\centering\hspace{0pt}}m{0.120\linewidth}} 
\toprule
\textbf{Related Works}         & \textbf{Algorithm} & \textbf{Single-Pass Learning} & \textbf{Online Learning \& Personalization} & \textbf{App Flexibility} & \textbf{Platform Agnostic}  \\ 
\hline
Rahimi et al.~\cite{rahimi2020hyperdimensional} & HDC & \xmark & \xmark   &  \xmark &  -    \\
Moin et al.~\cite{moin2021wearable}&  HDC  &  \cmark &  \xmark  &    \xmark  &  \xmark    \\
Benatti et al.~\cite{benatti2019online} &  HDC   &  \cmark & \xmark &   \xmark  &  \xmark    \\
Burrello et al.~\cite{burrello2020ensemble} & HDC &  \xmark    & \xmark &   \xmark   &  \xmark \\
Bhat et al.~\cite{bhat2019ultra} & DNN &  N/A    & N/A &   \xmark  &  \xmark  \\
\textbf{Ours} & \textbf{HDC} &  \textbf{\cmark}    & \textbf{\cmark} &  \cmark  &  \cmark \\
\bottomrule
\end{tabu}
}
\vspace{-1mm}
\label{tab:related}
\end{table*}
Almost every health monitoring system requires to meet the following criteria: \textit{\textbf{(a) Energy-efficiency}}: A practical e-health system needs to be efficient in using energy to enable long-term continuous monitoring on battery-powered wearable devices. \textbf{\textit{(b) Performance}}: it is vital to deliver real-time and accurate services for rapid response to emergency. \textbf{\textit{(c) Personalization}}: the system must use individual user information to achieve better intelligence while adapting to new user behavioral and physiological patterns over time. \textbf{\textit{(d) User privacy}}: To preserve privacy,  users are not often willing to transfer and store their private health information to the cloud \cite{ozanne2018wearables}. On-device processing can address this concern. Related health monitoring research using wearable devices has appeared in two contexts: offloading-based and on-device solutions, as summarized below.

\noindent\textbf{\textit{(a) Offloading-based Solutions:}} prior works propose solutions to transmit collected sensory data to an external device such as cloud servers. \cite{hussein2018automated,wan2018wearable} propose a cloud-based remote monitoring system for observing health status of the patients. Cloud-based solutions are conventional techniques to improve performance and efficiency. However, these solutions have some drawbacks. The monitoring systems must be reliable even when network connections are lost. In addition, transmitting user data to the cloud raises privacy concerns. Patel et al.~\cite{patel2016wearable} propose a platform to develop cloud-based ML models for health monitoring applications. However, the trained models are typically less personalized as they are trained to target a broad user community.

\noindent\textbf{\textit{(b) On-Device Solutions:}} several solutions \cite{preece2009activity,kirwan2012using} have implemented health monitoring services on smartphones, however, they all suffer from excessive energy usage as their power consumption is in the order of Watts. 
To address this challenge, some works have utilized intelligent gateways rather than smartphones for monitoring systems \cite{attal2015physical,islam2018design}. 
These solutions employ powerful embedded processors for running ML workloads. 
Other efforts have attempted to design custom hardware accelerators to meet the energy budget requirement \cite{klinefelter201521}. For example, in \cite{wong20081v,luo201793muw}, the authors propose a custom System-on-Chip for vital sign monitoring. More recently, \cite{bhat2019ultra} presents an ultra-low power custom hardware for the activity recognition application.
Although these solutions offer low-power and real-time (inference) services, they fail to provide on-device learning because of their highly computational-intensive nature. 

Recent efforts have adopted HDC algorithms for health applications~\cite{rahimi2020hyperdimensional,rahimi2017hyperdimensional,benatti2019online,burrello2020ensemble}. These works targeted specific applications such as hand gesture recognition~\cite{moin2021wearable} or Seizure Detection applications~\cite{burrello2020ensemble} and are implemented using ASICs~\cite{moin2021wearable}. The literature lacks a comprehensive solution which is \textit{flexible} and \textit{platform-agnostic} to run a variety of health applications, in particular on multi-purpose wearables. Our solution presents a pipeline from data collection to monitoring output as a general solution for health monitoring systems. In addition, its platform-agnostic property enables it be deployed on variety of platforms including CPU and FPGA based on user-defined requirements. It should be noted that the existing HDC-based solutions for health applications ~\cite{moin2021wearable,benatti2019online} support \textit{Single-pass Learning}, where learning can be performed in a single iteration looking at each training data point. While this approach enables fast and real-time learning from a stream of data, it suffers from very weak classification accuracy~\cite{hernandez2021onlinehd}. This is in contrast to the ultimate goal of personalization which is to maximize model performance. Our work demonstrates that online learning is feasible on wearable devices. Table~\ref{tab:related} summarizes and compares our contributions against the state-of-the-art. 

\section{HyperDimensional Classification}
We present a robust and lightweight hyperdimensional classification. 
The first step in HDC is to encode data into a high-dimensional space. Then, HDC performs a learning task over encoder data by performing a single-pass training that generates a hypervector representing each class. The inference task can performed by checking the similarity of an encoded query to the class hypervector. 

\subsection{Hyperdimensional Encoding}
Let's assume $\vec{\mathcal{H}}_1$,  $\vec{\mathcal{H}}_2$ are two randomly generated hypervectors ($\vec{\mathcal{H}} \in \{-1,+1\}^D$) and $\delta (\vec{\mathcal{H}}_1, \vec{\mathcal{H}}_2) \approx 0$.
HDC is based on a set of primitives: 
\textbf{(1) Bundling:} is an addition of multiple hypervectors into a single hypervector, $\hypvec{R}= \hypvec{V}_1 + \hypvec{V}_2$, where $\hypvec{V}_2 \in \{0,1\}^D$ and $D$ is the dimension of the HDC space. 
Unlike original space where bundling act as an average operation, in high-dimensional space the addition is memorization function. 
\textbf{(2) Binding:} associates multiple orthogonal hypervectors (e.g., $\hypvec{V}_1$, $\hypvec{V}_2$) into a single hypervector ($\hypvec{R}=\hypvec{V}_1  *  \hypvec{V}_2$). 
The bound hypervector is a new object in HDC space which is orthogonal to all input hypervectors ($\delta (\hypvec{R}, \hypvec{V}_1) \simeq 0$ and $\delta (\hypvec{R}, \hypvec{V}_2) \simeq 0$). \textbf{(3) Permutation:} defined as a single rotational shift. 
The permuted hypervector will be nearly orthogonal to its original hypervector ($\delta (\hypvec{V}_1 \rho \hypvec{V}_1) \simeq 0$).   

\begin{figure}[t!]
    \centering
         \includegraphics[width=0.85\columnwidth]{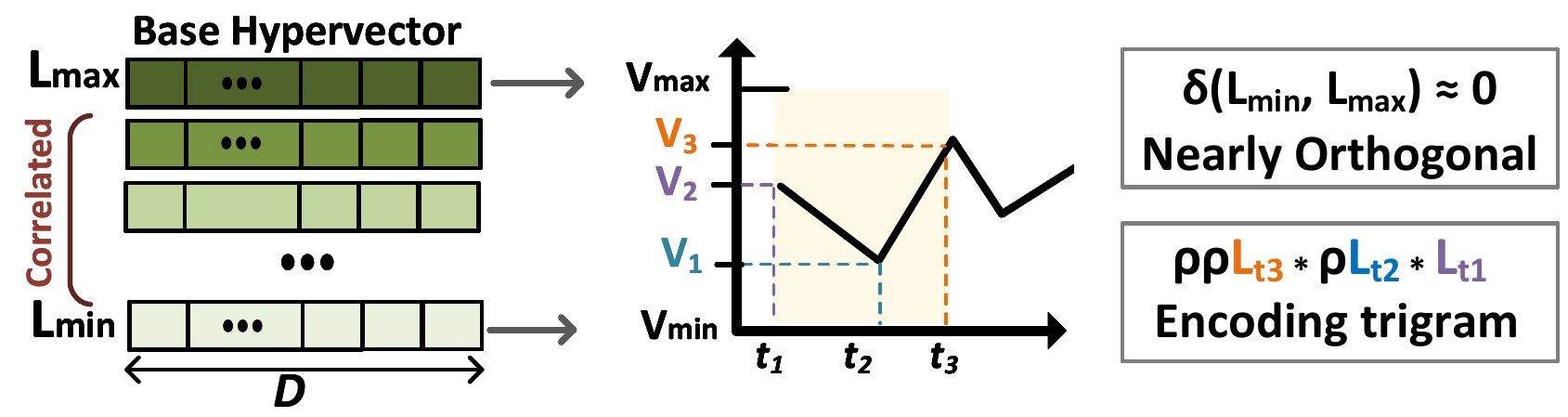}
    \vspace{-6mm}
    \caption{Time-series encoding to high-dimensional space.}
    \label{fig:encoding}
    \vspace{-3mm}
\end{figure} 

The encoding of the text data starts by assigning a random binary hypervector to each character in the alphabet. For example, for encoding  English text, we generate a random hypervector representing digits \texttt{A} to \texttt{Z}. We encode text data in a $n$-gram windows, where $n$ is usually a number between 3 to 5. Considering a trigram ``\texttt{A-B-C}'' example, we use the following embedding to map a sequence to high-dimensional space: $\rho\rho \vec{L}_A * \rho \vec{L}_B * \vec{L}_C$. The encoding module binds the hypervectors corresponding to alphabets while exploiting permutation to remember their sequence.  
HDC uses a very similar encoding as text data to map time-series into high-dimensional space. 
As Figure~\ref{fig:encoding} shows, we sample time-series in an $n$-gram window. In each sample window, the signal values (in the y-axis) stores the information, and the time (x-axis) represents the sequence. 
We assign a random vector to $V_{min}$ ($\vec{L}_{min}$ representing the minimum signal value) and $V_{max}$ ($\vec{L}_{max}$ representing the maximum signal value). 
Since these vectors are randomly generated, they are nearly orthogonal. 
For signal values between $V_{min}$ and $V_{max}$, we perform vector quantization to generate vectors that have a spectrum of similarity to $\vec{L}_{min}$ and $\vec{L}_{max}$ similarity. Finally, the encoding is performed by binding the level hypervectors corresponding to sampled signal while using permutation to store the timing information. 
For the example shown in Figure~\ref{fig:encoding}, the trigram Windows can be encoded as $\hypvec{H}= \rho\rho \vec{L}_{t_3} * \rho \vec{L}_{t_2} * \vec{L}_{t_1}$.
Note that for time-series, the encoding repeats after moving a sliding window one time-step ahead. This will generate large number of encoded data that can be used for training. 

For applications with multiple sensors, the encoding follows the same procedure for each sensor using synchronized $n$-gram windows. 
Let's assume a problem with $m$ sensors, where sensors generate the following encoded hypervectors $\{\hypvec{H}_1, \hypvec{H}_2, \cdots, \hypvec{H}_m\}$. 
We accordingly generate $m$ random hypervectors, where each is a signature of a sensor $\{\hypvec{P}_1, \hypvec{P}_2, \cdots, \hypvec{P}_m \}$. 
To aggregate information, the encoded hypervector from each sensor will be bound (associated) with the corresponding identification hypervector:

$$\hypvec{H} =\hypvec{P}_1 * \hypvec{H}_1+ \hypvec{P}_2 * \hypvec{H}_2 + \cdots + \hypvec{P}_m * \hypvec{H}_m $$

\noindent where `$+$' memorizes the sensor information and each $\hypvec{P}$ preserves the position of a sensor.

\subsection{Online Learning} \label{sec:singlepass}
We propose an adaptive training framework for efficient and accurate learning in HDC. 
Our training identifies common patterns during training and eliminates the 
 saturation of the class hypervectors during traditional single-pass training. 
Instead of naively combining all encoded data points, our approach adds each encoded data to class hypervectors depending on how much new information the pattern adds to class hypervectors. 
If a data point already exists in a class hypervector, \Design will add no or a tiny portion of data to the model to eliminate hypervector saturation. 
\begin{figure}[t!]
    \centering
         \includegraphics[width=0.85\columnwidth]{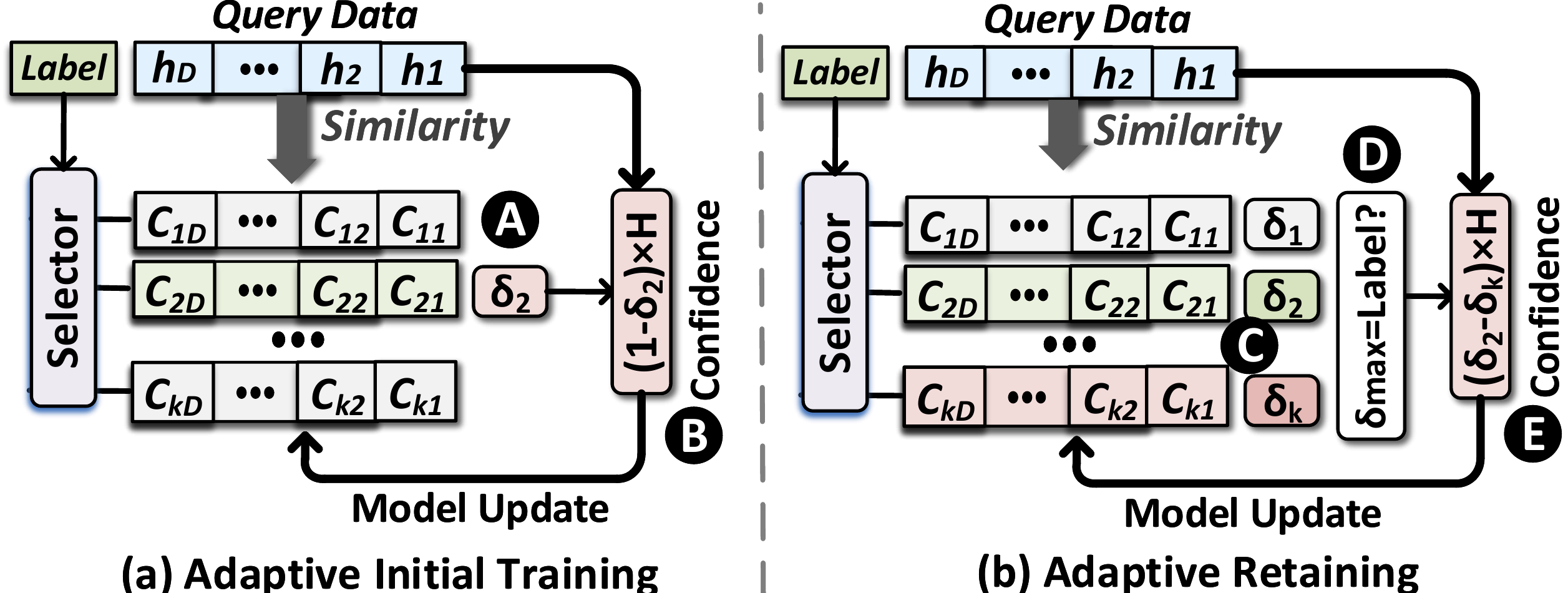}
    \vspace{-4mm}
    \caption{(a) HDC Online Learning (b) Iterative Learning.}
    \label{fig:overview}
    \vspace{-6mm}
\end{figure} 
Figure~\ref{fig:overview}a shows the framework's functionality during adaptive initial training. Let's assume $\vec{\mathcal{H}}$ as a new training data point. 
The framework computes the cosine similarity of $\vec{\mathcal{H}}$ with a class hypervector that has the same label as $\vec{\mathcal{H}}$. If the data point corresponds to $i^{th}$ class, we compute similarity of a data point with $\vec{\mathcal{C}}_i$ as: 
$\delta(\vec{\mathcal{H}}, \vec{\mathcal{C}^l}) = \frac{\vec{\mathcal{H}} \cdot \vec{\mathcal{C}^l}}{\mathbin{\parallel} \vec{\mathcal{H}} \mathbin{\parallel} \cdot \mathbin{\parallel} \vec{\mathcal{C}^l} \mathbin{\parallel}} $
where $\vec{\mathcal{H}} \cdot \vec{\mathcal{C}^l}$ is a dot product between a query and class hypervector ($\invcircledast{A}$).
The $\delta$ value shows the similarity of a data point to its class hypervector. Instead of naively adding data point to the model, \Design updates the model based on the $\delta$ similarity. 
For example, if an input data has label $l$, the model updates as follows ($\invcircledast{B}$).
\begin{equation}\label{eq:update}
\begin{split}
\vec{\mathcal{C}_l} \gets  \vec{\mathcal{C}_l} + \eta~(1-\delta_{l}) \times \mathcal{\vec{H}} 
\end{split}
\end{equation}
\noindent where $\eta$ is a learning rate. A large $\delta_{l}$ indicates that the input is a common data point which is already exist in the model. Therefore, our update adds a very small portion of encoded query to model to eliminate model saturation ($1-\delta_{l} \simeq 0$). However, small $\delta_{l}$ means that the query has new pattern which does not exist in the model. Thus, the model is updated with a larger factor ($1-\delta_{l} \simeq 1$). 

\subsection{Iterative Learning}
Although single-pass training is suitable for fast and ultra-efficient learning, embedded devices may have enough resources to perform more accurate learning tasks. 
Our framework supports retraining to enhance the quality of the model. 
Instead of starting to retrain from a naive initial model, \Design retraining starts from the initial adaptive model (explained in Section~\ref{sec:singlepass}). 
The framework's initial model already considered the weight of each input data during single-pass training. 
Therefore, \Design retraining starts from a well-trained initial model with relatively high classification accuracy. This enables \Design to retrain the model with a much lower number of iterations, resulting in fast convergence. 
Figure~\ref{fig:overview}b shows \Design functionality during adaptive retraining. 
\Design follows a similar learning procedure as initial training. For each training data point, say $H$, \Design checks the similarity of data with all class hypervectors in the model ($\invcircledast{C}$) and updates the model for each miss-prediction ($\invcircledast{D}$). 
Retraining examines if the model correctly returns the label $l$ for an encoded query $\vec{\mathcal{H}}$. If the model mispredicts it as label $l'$, the model updates as follows ($\invcircledast{E}$).
\begin{equation}\label{eq:update2}
\begin{split}
\vec{\mathcal{C}_l} \gets \vec{\mathcal{C}_l} + \eta~ (\delta_{l'}-\delta_{l}) \times \mathcal{\vec{H}} \\
\vec{\mathcal{C}_{l'}} \gets \vec{\mathcal{C}_{l'}} - \eta~ (\delta_{l'}-\delta_{l}) \times \mathcal{\vec{H}}
\end{split}
\end{equation}
\noindent where $\delta_l = \delta(H,\vec{\mathcal{C}_{l}})$ and $\delta_{l'} = \delta(H,\vec{\mathcal{C}_{l'}})$ are the similarity of data with correct and miss-predicted classes, respectively. This ensures that we update the model based on how far a train data point is miss-classified with the current model. In case of of a very far miss-prediction, $\delta_{l'}>>\delta_{l}$, \Design retraining makes a major changes on the mode. While in case of marginal miss-prediction, $\delta_{l'} \simeq	\delta_{l}$, the update makes smaller changes on the model. 
\section{HDC for Wearable Devices}
\subsection{System Architecture}
\begin{figure}[tb!]
    \centering
     \includegraphics[width=0.85\textwidth]{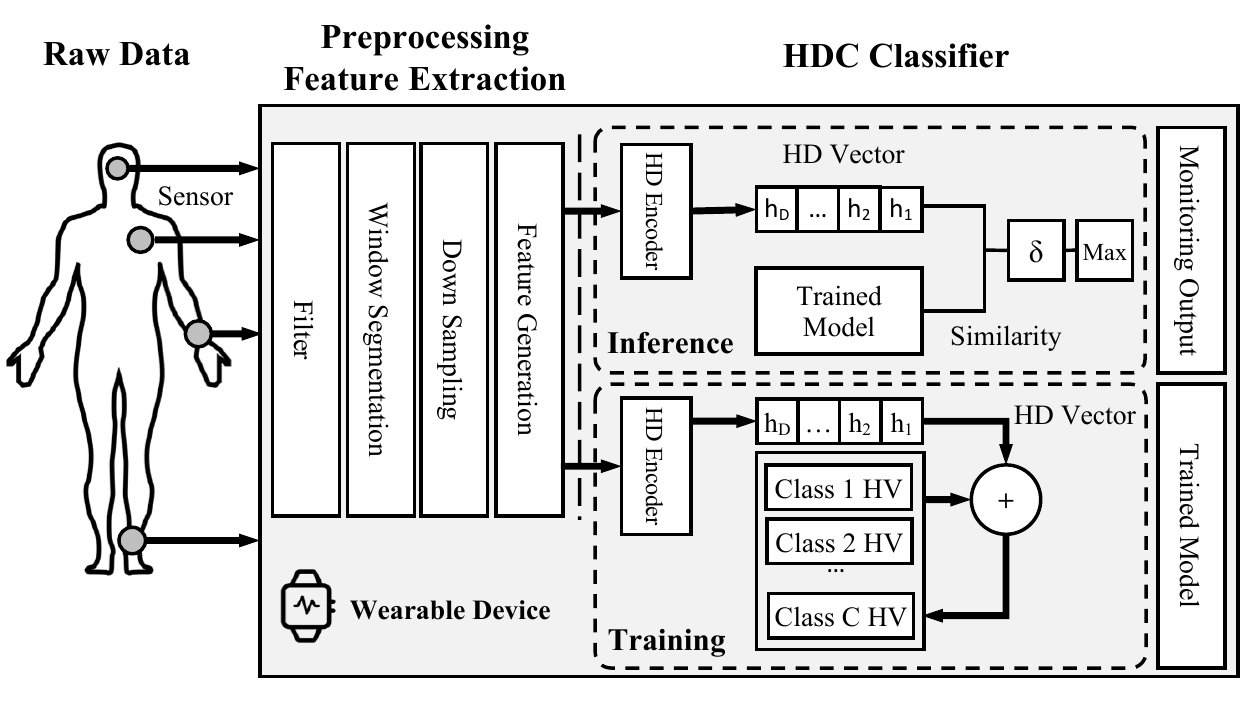}
      \caption{HDC for Health monitoring system architecture.}
      \label{fig:overview2}
      \vspace{-4mm}
    \end{figure}

Figure~\ref{fig:overview2} shows our proposed monitoring system. The proposed system collects the raw data from sensors such as electrocardiography (ECG), photoplethysmography (PPG), etc. Self-reported labels are collected through the wearable's user interface (e.g., an smartwatch's touchscreen) from subjects in-the-moment to personalize the model for each individual over time.
The collected sensory data is processed through two major steps at the wearable device: \textbf{(a)} \textbf{Preprocessing and Feature Extraction}, \textbf{(b)} \textbf{HDC Classifier}. The preparation includes data integration, data cleaning, and data reduction. Feature generation is the next step which derives values intended to be informative and non-redundant. It facilitates more accurate subsequent learning. The HDC classifier is the final step where the training and inference phase are performed. During the training, our framework provides single-shot and iterative training known as HD-Online and HD-Iterative, respectively. HD-Online identifies common patterns within a single-pass training. This results in learning the model as input data comes from the sensors. On the other hand, HD-Iterative offers retraining to enhance the quality of the model. 
In the following, we explain the health monitoring applications we use to evaluate our proposed system.
\subsection{Wearable Health Applications}
In the following, we describe our case studies for the health monitoring system:

\noindent\textbf{{(a) Pain Monitoring (PM):}} Pain is a single primary reason for people seeking medical care and is associated with many illnesses \cite{tompkins2017pain}. In acute pain management, pain assessment is critical for optimal pain treatment and evaluation for intervention decisions. However, pain, as a multivalent, dynamic, and ambiguous phenomenon is difficult to quantify \cite{raffaeli2017pain}. In particular, at times when a patient has limitations in his/her communication (e.g., during critical illness, infants and preverbal toddlers, patients under sedation or anesthesia, persons with intellectual disabilities, and patients at the end of life) \cite{breivik2008pain}. While pain is a highly subjective experience, behavioral and physiological manifestations of pain can be objectively measured. There have been efforts in developing objective pain assessment tools through analyzing changes in physiological pain indicators, such as heart rate (HR), heart rate variability (HRV), and electrodermal activity (EDA) \cite{solana2015pain,geuter2014pain}. 
In this work, we evaluate the pain monitoring system, which estimates the pain intensity using ECG signal from the BioVid dataset \cite{walter2013biovid}. 
We filter ECG signals using a band-pass filter with the frequency ranges of [0.1,250] Hz. This step is necessary to reduce noise and minimize the effects of trends in the signals. We extract three features from ECG signals and associate them with a particular pain level.

\noindent\textbf{(b) Stress Monitoring (SM):} Stress as an interesting affective state is defined as 'nonspecific response of the body to any demand upon it' \cite{wesad}. Long-term stress is known to have severe implications on well-being ranging from headaches and troubled sleeping to an increased risk of cardiovascular diseases \cite{chrousos1992concepts, mcewen1993stress, rosmond1998endocrine}. These severe side effects of stress call for automated stress monitoring systems. Monitoring people’s stress levels has become an essential part of behavioral studies for physical and mental illnesses. Building a reliable stress monitoring system requires understanding that it is primarily a physiological response to a stimulus triggered by the sympathetic nervous system (SNS). The current state of wearable sensor technology allows us to develop systems measuring the physiological signals reflecting stress 24/7 while capturing the context. We evaluate the Stress Monitoring system detecting the presence of stress from a dataset featuring PPG signals, recorded from a wrist-worn device, of 15 subjects during a lab study \cite{wesad}. We preprocess the PPG data using band-pass and moving average filters to remove noise from the data. Then, we extract $7$ features from the preprocessed PPG data.

\noindent\textbf{{(c) Human Activity Recognition (HAR)}:} Human activity recognition is the problem of classifying sequences of accelerometer data recorded by specialized harnesses or smartphones into known well-defined movements. The problem is to predict a user's movement based on sensor data such as an accelerometer in wearable devices. It is a challenging problem given the large number of observations produced each second, the temporal nature of the observations, and the lack of a straightforward way to relate accelerometer data to known movements. It involves predicting the movement of a person based on sensor data and traditionally involves deep domain expertise and methods from signal processing to correctly engineer features from the raw data to fit a machine learning model. Classical approaches to the problem involve hand crafting features from the time series data based on fixed-size windows and training machine learning models, such as for ensembles of decision trees. We evaluate the HAR system using an accelerometer and stretch-based dataset \cite{bhat2019ultra}. We preprocess the raw accelerometer data with a moving average filter. Then, we extract $7$ features from the filtered accelerometer data and segment them into activity windows.

\section{Evaluation and Analysis}
We implement the HD-based monitoring system on two embedded platforms: Raspberry Pi 3B+ using ARM CPU and Xilinx Kintex 7 FPGA. We implement HD functionality on FPGA platform using Verilog and synthesize it using Xilinx Vivado Design Suite based on a state-of-the-art FPGA framework~\cite{imani2021revisiting}. We demonstrate the effectiveness of our proposed system with three case studies, including HAR, PM, and SM applications (See Section IV.B). In the following, we evaluate our proposed system's accuracy, performance, and energy efficiency on these CPU and FPGA platforms.
We compare HD algorithm accuracy against state-of-the-art learning algorithms, including Deep Neural Network (DNN), Support Vector Machine (SVM). DNN and SVM models are trained with Tensorflow and the Scikit-learn library, respectively. We use the common practice of the grid search to identify the best hyperparameters for each model. The neural network architecture consists two hidden layers with $512$, and $128$ neurons. The HD algorithm is trained using a single-pass (HD-Online) and iterative (HD-Iterative) way using $D=4k$. 
\begin{figure}[t!]
\pgfplotsset{every x tick label/.append style={font=\tiny, yshift=0.5ex}}
\pgfplotsset{every y tick label/.append style={font=\tiny, xshift=0.1ex}}
\usetikzlibrary{patterns}
\pgfplotsset{compat=1.11,
	/pgfplots/ybar legend/.style={
		/pgfplots/legend image code/.code={%
			\draw[##1,/tikz/.cd,yshift=-1mm]
			(5mm,2mm) rectangle (2pt,0.2em);},
	},
}
\centering
\begin{tikzpicture}
\begin{axis}[
ybar=2pt,
grid=major,
enlarge x limits={abs=0.6},
ymin=0,
width  = 9cm,
height = 4.5cm,
bar width=8pt,
ylabel={\normalsize Accuracy (\%)},
xticklabel style={rotate=0, font=\small},
yticklabel style={rotate=0, font=\normalsize},
xtick = data,
ylabel near ticks,
table/header=false,
every node near coord/.append style={font=\tiny},
table/row sep=\\,
xticklabels from table={
	HAR\\
	PM\\
	SM\\
}{[index]0},
legend columns=4,
enlarge y limits={value=.05,upper},
legend style={at={(0.5,-.2)},anchor=north, font=\normalsize},
]
\legend{HD-Iterative,HD-Online, SVM, DNN}

\addplot [draw=black, pattern=horizontal lines] table[x expr=\coordindex,y index=0]{85.79\\98.33\\85.40\\};
\addplot [draw=black,fill=gray]  table[x expr=\coordindex,y index=0,red]{77.2\\86.35\\77.2\\};
\addplot [draw=black,pattern=north east lines] table[x expr=\coordindex,y index=0]{81.35\\97.34\\78.2\\};
\addplot [draw=black,fill=black] table[x expr=\coordindex,y index=0]{88.1\\95.20\\77.05\\};

\pgfplotsinvokeforeach{0,1,2,3}{\coordinate(l#1)at(axis cs:#1,0);}
\end{axis}
\end{tikzpicture}
\caption{HD vs state-of-the-art accuracy analysis.} 
\vspace{-4mm}
\label{fig:accuracy}
\end{figure}
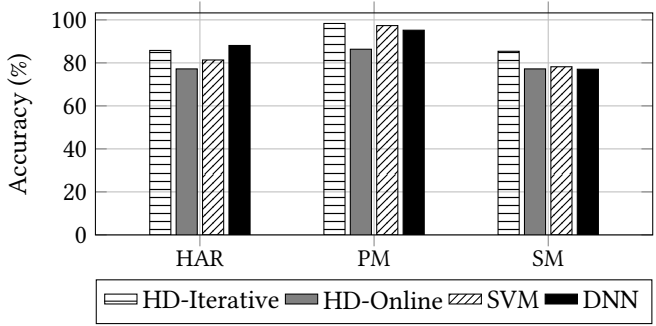

\subsection{Accuracy Analysis}
In this subsection, we demonstrate the prediction accuracy of the HD algorithm through experimental comparisons against state-of-the-art learning algorithms. We demonstrate how personalization can be achieved using the HDC-based classifiers. We first train the models based on the collected data from all subjects. Figure \ref{fig:accuracy} shows that the HD approach provides comparable accuracy to the state-of-the-art learning algorithms for three health monitoring applications. We report the results for HD algorithm for both iterative and online strategies. The HD-Iterative results show errors of $0.03\%$ and $0.02\%$ in comparison with DNN the algorithm for the HAR and Stress Monitoring applications, respectively. However, HD-iterative is even $0.1\%$ more accurate than DNN algorithm for the PM application. On the other hand, the HD-Online method leads to $6.9\%$, $8.8\%$, and $2.7\%$ accuracy degradation compared with the DNN algorithm for HAR, PM, and SM, respectively. Our evaluation shows HD-iterative leads to $4.4\%$, $0.1\%$ accuracy improvement compared with SVM learning algorithm for HAR and PM application, respectively. While, HD-Online results in $2\%$, $8\%$ accuracy error in comparison with SVM learning algorithm.

\begin{table}
\caption{Accuracy Analysis for the Personalized model vs. the General model for the Stress monitoring application. Pers. and Gen. represent the Personalized and General models. S1 to S6 represent six participants in the study.}
\label{tab:personal}
\resizebox{0.75\textwidth}{!}{
\begin{tabular}{@{}clllllll@{}}
\toprule
\textbf{Model} &
  \multicolumn{1}{c}{\textbf{Strategy}} &
  \multicolumn{1}{c}{\textbf{S1}} &
  \multicolumn{1}{c}{\textbf{S2}} &
  \multicolumn{1}{c}{\textbf{S3}} &
  \multicolumn{1}{c}{\textbf{S4}} &
  \multicolumn{1}{c}{\textbf{S5}} &
  \multicolumn{1}{c}{\textbf{S6}} \\ \midrule
\multirow{2}{*}{\textbf{Pers.}} & \textbf{Online}    & 77.2 \% & 78.2\% & 79.2\% & 64.6\% & 68.5\% & 64.9\% \\
                                       & \textbf{Iterative} & 85.4\% & 87.4\% & 88.2\% & 79.6\% & 82.9\% & 79.6\% \\ \midrule
\multirow{2}{*}{\textbf{Gen.}}      & \textbf{Online}  & 50.6\%  &  49.9\%  &  51.8\% &  48.3\%  &  50.1\%  & 59.0\% \\
                                       & \textbf{Iterative} &  72.6\%      &    62.9\%    &     73.6\%   &  74.5\%  &  74.6\%    &  76.6\%      \\ \bottomrule
\end{tabular}
}
\vspace{-3mm}
\end{table}

The bias in physiological data can be different for personal or general dataset~\cite{han2020objective}. We report the effect of personalization and how it improves the monitoring accuracy. We evaluate the personalization considering six participants (S1-S6) for the Stress monitoring application. To train the \textit{General} model, we exclude the data from one subject and then train the model using data from all other subjects. We test the model on half of the data from the excluded subject (selected randomly). To train the \textit{Personalized} model, we use the first half of each subject's data for training (to emulate the progression of time) and then test it with the second half of the subject's data. Table~\ref{tab:personal} shows \textit{Personalized} model performance in comparison with \textit{General} model. The results show an average of 20.48\% and 11.38\% improvement on classification accuracy when personalization is used for both HD-Online and HD-Iterative, respectively.

\subsection{Performance Analysis}
\begin{figure}[tb!]
\pgfplotsset{every x tick label/.append style={font=\tiny, yshift=0.5ex}}
\pgfplotsset{every y tick label/.append style={font=\tiny, xshift=0.5ex}}
\usetikzlibrary{patterns}
\pgfplotsset{compat=1.11,
	/pgfplots/ybar legend/.style={
		/pgfplots/legend image code/.code={%
			\draw[##1,/tikz/.cd,yshift=-1mm]
			(5mm,2mm) rectangle (2pt,0.2em);},
	},
}
\begin{tikzpicture}
\begin{axis}[
ybar=4pt,
grid=major,
enlarge x limits={abs=0.6},
ymin=0,
width  = 9cm,
height = 4.5cm,
bar width=8pt,
ylabel={\normalsize Speedup (DNN=1)},
xticklabel style={rotate=0, font=\small},
yticklabel style={rotate=0, font=\normalsize},
xtick = data,
ylabel near ticks,
table/header=false,
every node near coord/.append style={font=\tiny},
table/row sep=\\,
xticklabels from table={
	HAR\\
	PM\\
	SM\\
}{[index]0},
legend columns=4,
enlarge y limits={value=.1,upper},
legend style={at={(0.5,-0.2)},anchor=north, font=\normalsize},
]
\legend{HD-Iterative,HD-Online, SVM, DNN}
\addplot [draw=black,pattern=horizontal lines] table[x expr=\coordindex,y index=0]{4.87\\16.66\\16.66\\}; 
\addplot [draw=black,fill=gray]  table[x expr=\coordindex,y index=0,red]{14.9\\37.03\\25\\};
\addplot [draw=black,pattern=north east lines] table[x expr=\coordindex,y index=0]{9.61\\1.86\\8.13\\};
\pgfplotsinvokeforeach{0,1,2,3}{\coordinate(l#1)at(axis cs:#1,0);}
\end{axis}
\end{tikzpicture}
\caption{{Training time for HD vs. state-of-the-art on CPU platform.}}
\label{fig:perf}
\end{figure}
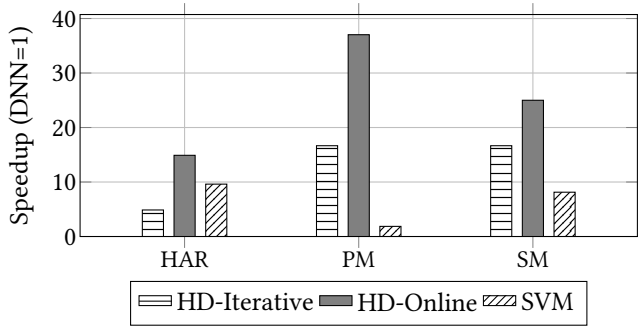

We report the performance of the HD learning algorithm during the training and inference phase on the platforms mentioned above. We evaluate three health monitoring applications, including HAR, PM, and SM, with HD and state-of-the-art learning algorithms. Figure \ref{fig:perf} shows that HD algorithm provides significant speedup for training time compared with other algorithms. HD-Iterative algorithm significantly reduces training time by $4.8\times$, $15.3\times$, and $15.8\times$ compared with the DNN algorithm for HAR, PM and SM applications, respectively.
On the other hand, HD-Online algorithm results in better training time where it provides $14.6\times$, $35.8\times$, and $23.81\times$ compared with DNN algorithm for HAR, PM, and SM applications, respectively. This speedup comes from HD-Online algorithm capability in lowering number of required training iterations. Table \ref{tab:perf} also compares HD inference and training phase performance with the state-of-the-art learning algorithms. HD algorithm presents $32\times$, $43.5\times$, and $4.29\times$ improvement in inference time compared with SVM learning algorithm for HAR, PM, and SM, respectively. 

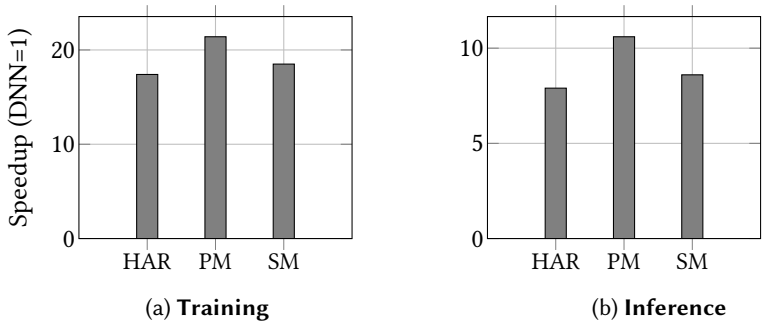
\begin{figure}[t!]
\pgfplotsset{every x tick label/.append style={font=\tiny, yshift=0.5ex}}
\pgfplotsset{every y tick label/.append style={font=\tiny, xshift=0.5ex}}
\usetikzlibrary{patterns}
\pgfplotsset{compat=1.11,
	/pgfplots/ybar legend/.style={
		/pgfplots/legend image code/.code={%
			\draw[##1,/tikz/.cd,yshift=-1mm]
			(5mm,2mm) rectangle (2pt,0.2em);},
	},
}
\begin{subfigure}{0.4\textwidth}
\begin{tikzpicture}
\begin{axis}[
ybar=4pt,
grid=major,
enlarge x limits={abs=1},
ymin=0,
height = 4.5cm,
bar width=8pt,
ylabel={\normalsize Speedup (DNN=1)},
xticklabel style={rotate=0, font=\small},
yticklabel style={rotate=0, font=\normalsize},
xtick = data,
ylabel near ticks,
table/header=false,
every node near coord/.append style={font=\tiny},
table/row sep=\\,
xticklabels from table={
	HAR\\
	PM\\
	SM\\
}{[index]0},
legend columns=4,
enlarge y limits={value=.1,upper},
legend style={at={(0.5,-0.2)},anchor=north, font=\normalsize},
]
\addplot [draw=black,draw=black,fill=gray] table[x expr=\coordindex,y index=0]{17.4\\21.4\\18.5\\}; 
\pgfplotsinvokeforeach{0,1,2,3}{\coordinate(l#1)at(axis cs:#1,0);}
\end{axis}
\end{tikzpicture}
\caption{\textbf{Training}}
\end{subfigure}
\hspace{2.7mm}
\begin{subfigure}{0.4\textwidth}
\begin{tikzpicture}
\begin{axis}[
ybar=4pt,
grid=major,
enlarge x limits={abs=1},
ymin=0,
height = 4.5cm,
bar width=8pt,
xticklabel style={rotate=0, font=\small},
yticklabel style={rotate=0, font=\normalsize},
xtick = data,
ylabel near ticks,
table/header=false,
every node near coord/.append style={font=\tiny},
table/row sep=\\,
xticklabels from table={
	HAR\\
	PM\\
	SM\\
}{[index]0},
legend columns=4,
enlarge y limits={value=.1,upper},
legend style={at={(0.5,-0.2)},anchor=north, font=\normalsize},
]
\addplot [draw=black,fill=gray] table[x expr=\coordindex,y index=0]{7.9\\10.6\\8.6\\}; 
\pgfplotsinvokeforeach{0,1,2,3}{\coordinate(l#1)at(axis cs:#1,0);}
\end{axis}
\end{tikzpicture}
\caption{\textbf{Inference}}
\end{subfigure}
\vspace{-2mm}
\caption{{Performance of HD-iterative vs. DNN on FPGA platform.}}
\label{fig:perffpga}
\vspace{-4mm}
\end{figure}

In addition, Figure \ref{fig:perffpga} shows performance evaluation for HD algorithm during training and inference on the FPGA platform. HD algorithm can achieve up to $21.4\times$ and $10.6\times$ speedup during training and inference, respectively. 
Comparing the results between the FPGA and CPU platforms shows the FPGA design significantly improves the performance mainly because CPUs use the same number of resources to perform 1-bit or 8-bit arithmetic operations, which limits the degree of parallelism in the CPU~\cite{imani2021revisiting}. 
In contrast, FPGAs are significantly efficient for implementing low-precision arithmetic operations~\cite{imani2021revisiting}. 
The FPGA design can perform parallel bitwise operations and enable fast and efficient HD computation.

\begin{table}[b]
\caption{Performance analysis HD vs. state-of-the-art on CPU platform.}
\label{tab:perf}
\resizebox{0.7\textwidth}{!}{
\begin{tabular}{@{}cccccc@{}}
\multicolumn{1}{l}{}                                  & \multicolumn{1}{l}{} & \multicolumn{4}{c}{\textbf{Execution Time (sec)}} \\
 &  & \multicolumn{1}{c|}{\textbf{HD-Iterative}} & \multicolumn{1}{c|}{\textbf{HD-Online}} & \multicolumn{1}{c|}{\textbf{SVM}} & \textbf{DNN} \\ \midrule
\multicolumn{1}{c|}{\multirow{2}{*}{\textbf{HAR}}}    & Training             & 3.10       & 1.02       & 1.58       & 15.10      \\
\multicolumn{1}{c|}{}                                 & Inference            & 0.01       & 0.01       & 0.32       & 0.01       \\ \midrule
\multicolumn{1}{c|}{\multirow{2}{*}{\textbf{PM}}}   & Training             & 3.59       & 1.54       & 29.67      & 55.25      \\
\multicolumn{1}{c|}{}                                 & Inference            & 0.08       & 0.08       & 3.48       & 0.35       \\ \midrule
\multicolumn{1}{c|}{\multirow{2}{*}{\textbf{SM}}} & Training             & 1.89       & 1.26       & 3.74       & 30.17      \\
\multicolumn{1}{c|}{}                                 & Inference            & 0.97       & 0.98       & 4.21       & 0.45       \\ \bottomrule
\end{tabular}
}
\end{table}

\subsection{Energy Analysis}
\begin{figure}[t!]
\pgfplotsset{every x tick label/.append style={font=\tiny, yshift=0.5ex}}
\pgfplotsset{every y tick label/.append style={font=\tiny, xshift=0.5ex}}
\usetikzlibrary{patterns}
\pgfplotsset{compat=1.11,
	/pgfplots/ybar legend/.style={
		/pgfplots/legend image code/.code={%
			\draw[##1,/tikz/.cd,yshift=-1mm]
			(5mm,2mm) rectangle (2pt,0.2em);},
	},
}
\begin{subfigure}{0.4\textwidth}
\begin{tikzpicture}
\begin{axis}[
ybar=4pt,
grid=major,
enlarge x limits={abs=1},
ymin=0,
height = 4.5cm,
bar width=8pt,
ylabel={\normalsize Energy Eff (DNN=1)},
xticklabel style={rotate=0, font=\small},
yticklabel style={rotate=0, font=\normalsize},
xtick = data,
ylabel near ticks,
table/header=false,
every node near coord/.append style={font=\tiny},
table/row sep=\\,
xticklabels from table={
	HAR\\
	PM\\
	SM\\
}{[index]0},
legend columns=4,
enlarge y limits={value=.1,upper},
legend style={at={(0.5,-0.2)},anchor=north, font=\normalsize},
]
\addplot [draw=black,draw=black,fill=gray] table[x expr=\coordindex,y index=0]{35.9\\45.8\\41.7\\}; 
\pgfplotsinvokeforeach{0,1,2,3}{\coordinate(l#1)at(axis cs:#1,0);}
\end{axis}
\end{tikzpicture}
\caption{\textbf{Training}}
\end{subfigure}
\hspace{3mm}
\begin{subfigure}{0.4\textwidth}
\begin{tikzpicture}
\begin{axis}[
ybar=4pt,
grid=major,
enlarge x limits={abs=1},
ymin=0,
height = 4.5cm,
bar width=8pt,
xticklabel style={rotate=0, font=\small},
yticklabel style={rotate=0, font=\normalsize},
xtick = data,
ylabel near ticks,
table/header=false,
every node near coord/.append style={font=\tiny},
table/row sep=\\,
xticklabels from table={
	HAR\\
	PM\\
	SM\\
}{[index]0},
legend columns=4,
enlarge y limits={value=.1,upper},
legend style={at={(0.5,-0.2)},anchor=north, font=\normalsize},
]
\addplot [draw=black,fill=gray] table[x expr=\coordindex,y index=0]{3.6\\4.8\\5.1\\}; 
\pgfplotsinvokeforeach{0,1,2,3}{\coordinate(l#1)at(axis cs:#1,0);}
\end{axis}
\end{tikzpicture}
\caption{\textbf{Inference}}
\end{subfigure}
\vspace{-2mm}
\caption{{Energy efficiency of HD-iterative vs. DNN on FPGA platform.}}
\label{fig:energyfpga}
\vspace{-4mm}
\end{figure}
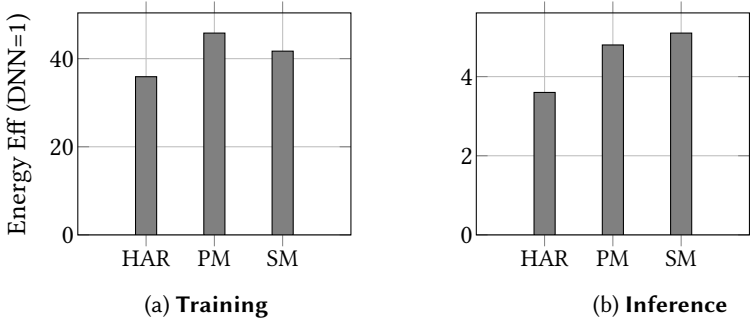

We compare energy efficiency for HD and DNN algorithms during training and inference on both FPGA and CPU platforms. Figure \ref{fig:energyfpga} shows that the HD algorithm provides a significant energy efficiency improvement during training and inference. The results are respective to DNN implementation on the FPGA platform. The energy efficiency improvement comes from eliminating costly gradient operations utilizing HD algorithm. In addition, HD algorithm uses simple bit-wise operations, which execute faster utilizing FPGA lookup tables (LUTs). Our evaluation of the health monitoring applications shows HD algorithm can improve energy efficiency during training and inference up to $45.8\times$ and $5.1\times$ on FPGA platform, respectively. The result shows that energy efficiency on CPU platform follows the performance trend during training and inference (See Figure \ref{fig:perf}).

\subsection{Robustness Analysis}
Table \ref{tab:robust} shows HD and DNN computation robustness for different health applications. For HD algorithm, the results are reported with $1$ bit representation and $4k$ dimension, which provide maximum accuracy. In DNN algorithm, weights are represented as $8$ bits values. In order to evaluate robustness to the noise in hardware, we randomly flip a certain amount of bits. The evaluation shows that HD algorithm provides significantly higher robustness to hardware error in comparison with DNN algorithm. In HD algorithm, all dimensions are equally contributing to stored information. Therefore, any failure on data only fails a portion of each hypervector, not losing the entire information. However, in DNN algorithm, any failure in stored weights leads to a significant classification accuracy drop. Our evaluation shows that HD algorithm is up to $23.1\times$, $60.1\times$, and $22.9\times$, more robust than DNN algorithm for HAR, SM, and PM applications, respectively. 

\begin{table}
\caption{Accuracy loss using noisy hardware for HD vs state-of-the-art DNN algorithm.}
\label{tab:robust}
\resizebox{0.65\textwidth}{!}{
\begin{tabular}{lcccccccc}
                      & \multicolumn{1}{l}{} & \multicolumn{1}{l}{} & \multicolumn{6}{c}{\textbf{Hardware Error}} \\
 &
  &
  &
  \multicolumn{1}{c|}{\textbf{1\%}} &
  \multicolumn{1}{c|}{\textbf{2\%}} &
  \multicolumn{1}{c|}{\textbf{4\%}} &
  \multicolumn{1}{c|}{\textbf{6\%}} &
  \multicolumn{1}{l|}{\textbf{10\%}} &
  \multicolumn{1}{l}{\textbf{12\%}} \\ \hline
\multicolumn{1}{l|}{} & \multicolumn{1}{c|}{}                                               & HD          & 0     & 0      & 0      & 0.3    & 0.5   & 0.8   \\
\multicolumn{1}{l|}{} & \multicolumn{1}{c|}{\multirow{-2}{*}{\textbf{HAR}}}                 & DNN         & 0.7   & 3.2    & 5.2    & 8.5    & 12.9  & 18.5  \\ \cline{2-9} 
\multicolumn{1}{l|}{} & \multicolumn{1}{c|}{}                                               & HD        & 0     & 0      & 0.1    & 0.4    & 0.7   & 0.8   \\
\multicolumn{1}{l|}{} & \multicolumn{1}{c|}{\multirow{-2}{*}{\textbf{SM}}} & DNN & 3.4   & 10.2 & 19.6 & 22.3   & 29    & 48.1  \\ \cline{2-9} 
\multicolumn{1}{l|}{} & \multicolumn{1}{c|}{} & HD          & 0     & 0      & 0.1    & 0.5    & 1.1   & 1.6   \\
\multicolumn{1}{l|}{\parbox[t]{5mm}{\multirow{-6}{*}{\rotatebox[origin=c]{90}{\textbf{Acc Error (\%)}}}}} &
\multicolumn{1}{c|}{\multirow{-2}{*}{\textbf{PM}}} &
DNN &
  4.5 &
  6.4 &
  12.4 &
  17.9 &
  \multicolumn{1}{l}{25.9} &
  \multicolumn{1}{l}{36.7} \\ \hline
\end{tabular}
}
\end{table}

\section*{Acknowledgements}
This work was supported in part by National Science Foundation (NSF) \#2127780, Semiconductor Research Corporation (SRC) Task \#2988.001, Department of the Navy, Office of Naval Research, grant \#N00014-21-1-2225 and \#N00014-22-1-2067, Air Force Office of Scientific Research, and a gift from Cisco.

\section{Conclusions}
We proposed an adaptive HDC training framework for health monitoring systems
that achieves fast, energy-efficient, and accurate on-device training/inference, and also enables personalization and privacy protection for wearable devices. We demonstrated the efficacy of our HDC approach using three realistic wearable healthcare studies,
achieving better energy  efficiency  for  training  and  inference  by up  to 45.8× and 5.1× compared  to state-of-the-art DNN  algorithms while achieving comparable  accuracy.
We believe that our HDC-based learning framework is a promising approach to meet the low-power, personalization, and privacy requirements for wearable health monitoring applications.

\bibliographystyle{unsrt}
\bibliography{related}
\end{document}